\documentclass[conference]{IEEEtran}
\IEEEoverridecommandlockouts
\usepackage{cite}
\usepackage{amsmath,amssymb,amsfonts}
\usepackage{graphicx}
\usepackage{booktabs}
\usepackage{tikz}
\usetikzlibrary{arrows.meta, positioning}
\usepackage{hyperref}
\hypersetup{hidelinks}

\begin{document}

\title{RiLM: Parameter-Efficient Language Modeling via Geodesic Decoding}

\author{\IEEEauthorblockN{Fang Li}
\IEEEauthorblockA{Department of Computer Science\\
    Oklahoma Christian University\\
    Edmond, USA\\
    Email: fang.li@oc.edu}
}

\maketitle

\begin{abstract}
Language models under one million parameters matter for edge deployment, domain adaptation, and reproducible research, yet a two-layer LSTM or Transformer at embedding width $d{=}128$ still spends roughly one third of its capacity on the output matrix $W_{\mathrm{out}} \in \mathbb{R}^{d \times |V|}$. We propose \emph{Riemannian Language Models} (RiLM), which remove that layer entirely: context unfolds as a trajectory on a Riemannian manifold, and next-token probabilities arise from squared geodesic distance between the current state and vocabulary embeddings. The same embedding map serves input and output---decoding \emph{is} geometry. We instantiate the framework on flat $\mathbb{R}^d$ (Flat RiLM) and the Poincar\'e ball $\mathbb{H}^d$ (HypRiLM) with a shared MLP composition map $\varphi$ (${\sim}290$k parameters, $d{=}128$, $|V|{=}2000$). Across five seeds on WikiText-2, HypRiLM reaches $54.2 \pm 0.2$ validation perplexity versus $87.6 \pm 0.6$ for Flat RiLM; tied and matched LSTM, Transformer, and SSM controls remain at $113$--$147$ PPL on WT-2---HypRiLM leads by roughly $2\times$ over the strongest tied recurrent baseline (SSM, $113.0 \pm 3.8$). Penn Treebank and a 10k-vocabulary stress test confirm that geodesic decoding transfers across corpora and larger $|V|$, while hyperbolic curvature helps selectively. We also characterize \emph{boundary collapse} in naive hyperbolic recurrence and show how M\"obius stabilization restores trainability. Claims are scoped to controlled small-model comparisons, not full-vocabulary state of the art.
\end{abstract}

\begin{IEEEkeywords}
language modeling, Riemannian geometry, hyperbolic neural networks, parameter efficiency
\end{IEEEkeywords}

\section{Introduction}
\label{sec:intro}

Neural language models map token sequences to distributions over the next word. The dominant recipe---token vectors in $\mathbb{R}^d$, state updates via LSTM gating~\cite{hochreiter1997} or self-attention~\cite{vaswani2017}, and an affine readout $W_{\mathrm{out}} h + b$---scales to billions of parameters and underpins systems from machine translation to code generation~\cite{radford2019,dai2019}. That recipe is the right default when compute and data are abundant. It is a poor default when the model must fit on device, adapt quickly to a narrow domain, or serve as a \emph{reproducible} scientific instrument: below one million parameters, $W_{\mathrm{out}} \in \mathbb{R}^{d \times |V|}$ can consume one third of the budget at $d{=}128$ and $|V|{=}2000$, matching the embedding table. Shrinking $d$ to compensate weakens the recurrent or attention core; enlarging $|V|$ without enlarging the model makes the output layer dominate entirely. Tied embeddings~\cite{press2017} halve redundant rows but still decode via a linear functional detached from how states move through representation space.

Language also carries hierarchical structure---syntax, lexical relations, topical organization---that Euclidean space embeds with distortion that grows in depth~\cite{nickel2017}. Hyperbolic space offers exponentially growing volume and has improved static embeddings and feed-forward layers~\cite{ganea2018}, yet it is rarely used to govern \emph{how} context is composed step by step. We ask whether manifold geometry can replace not only word vectors but the decoder itself.

We introduce \emph{Riemannian Language Models} (RiLM). The state $h_t$ lives on a Riemannian manifold $(M,g)$. A single map $\varphi$, shared at every timestep, reads the current state and incoming token and proposes a tangent direction; the state then moves along the corresponding geodesic. Decoding needs no $W_{\mathrm{out}}$: word $w$ is likely when its embedding $e_w$ lies close to $h_t$ in geodesic distance,
$p(w \mid h_t) \propto \exp(-d_M(h_t,e_w)^2/\tau)$.
Context becomes a trajectory $h_0 \!\to\! h_1 \!\to\! \cdots \!\to\! h_T$; classification and representation are coupled because the model must arrange vocabulary points so that plausible continuations lie near that trajectory.

We study two instantiations: Flat RiLM on $M{=}\mathbb{R}^d$ (the Euclidean special case of our earlier unpublished prototype) and HypRiLM on the Poincar\'e ball $M{=}\mathbb{H}^d_c$. Hyperbolic recurrence posed an unexpected obstacle---naive exponential-map updates drive states to the ball boundary and flatten all logits---which we analyze and resolve with M\"obius translation (Section~\ref{sec:discussion}). Empirically, we report multi-seed WikiText-2 and Penn Treebank results against untied, tied, and matched-parameter LSTM, Transformer, and selective state-space baselines. HypRiLM improves ${\sim}38\%$ relative to Flat RiLM on web text and trails Flat RiLM on newswire PTB; on both corpora, each RiLM variant beats matched and tied LSTM, Transformer, and SSM controls by wide margins (Tables~\ref{tab:fair} and~\ref{tab:ptb}). We do not target billion-parameter pretraining; the contribution is a geometric language-modeling template, a trainability analysis for hyperbolic recurrence, and evidence that decoding geometry matters when parameters are scarce.

\textbf{Contributions.} This paper makes four claims, each backed by controlled experiments at ${\sim}290$k parameters. First, \emph{geodesic decoding} replaces $W_{\mathrm{out}}$ with distance to vocabulary points on $M$, coupling representation and classification without a separate readout layer. Second, the same template instantiates flat and hyperbolic sequential models, enabling a direct test of whether curvature helps composition when $\varphi$ is held fixed. Third, we identify \emph{boundary collapse} in naive hyperbolic recurrence---a failure mode that drives perplexity to the uniform $|V|$ baseline---and show that M\"obius-based updates restore stable training. Fourth, under explicit fairness protocols (tied and matched baselines, including a modern SSM control), both RiLM variants outperform LSTM, Transformer, and SSM decoders on WikiText-2 and Penn Treebank by margins that tied embeddings alone do not explain.

\textbf{Scope and organization.} Our primary evaluation uses the 2000 most frequent words on WikiText-2 and PTB so that $W_{\mathrm{out}}$ accounts for a transparent ${\sim}256$k parameters at $d{=}128$; we do not claim numerically comparable perplexity to full-vocabulary leaderboard models. Section~\ref{sec:method} defines RiLM; Section~\ref{sec:experiments} presents multi-seed results, vocabulary scaling, ablations, and geometry analysis; Section~\ref{sec:discussion} interprets boundary collapse, when curvature helps, and limitations.

\section{Related Work}
\label{sec:related}

\textbf{Recurrent and attention decoders.} Modern LMs stack depth, width, and data until perplexity on WikiText-2 drops into the twenties---but only at 33M--257M parameters with heavy regularization~\cite{merity2018,dai2019,radford2019}. At sub-million scale the bottleneck shifts: a two-layer LSTM or Transformer must either carry a full $W_{\mathrm{out}}$ or tie embeddings~\cite{press2017}. Tying removes redundant rows but preserves the \emph{form} of decoding as a linear functional on hidden states in $\mathbb{R}^d$. RiLM instead decodes by geometry on $M$: the hidden state and vocabulary live in the same space, and logits are derived from distances. We compare against both untied models (to show the cost of $W_{\mathrm{out}}$) and tied/matched models (to isolate the effect of decoding geometry).

\textbf{Hyperbolic representation learning.} Hyperbolic space embeds trees and hierarchies with low distortion~\cite{nickel2017}; hyperbolic neural networks provide M\"obius operations and Riemannian optimization tools~\cite{ganea2018}. Most prior work applies hyperbolicity to \emph{static} objects---word vectors, graph nodes, or single feed-forward layers. RiLM asks whether curvature should govern the \emph{trajectory} of context: each token applies the same $\varphi$ and moves the state along a geodesic, and every prediction step reuses the same distance-based readout. The sequential, recurrent use of hyperbolic geometry is therefore qualitatively different from Poincar\'e word embeddings with a Euclidean softmax on top.

\textbf{Efficiency and modern recurrent baselines.} Parameter-efficient fine-tuning (e.g., LoRA~\cite{hu2022}) reduces adaptation cost but leaves the pretrained decoder intact. Continuous-time models~\cite{chen2018} and volume-preserving flows~\cite{rezende2015} study dynamics but do not remove vocabulary-sized output layers in small LMs. Selective state-space models~\cite{gu2023mamba} offer linear-time recurrence with input-dependent gating and represent the strongest modern alternative to LSTM/Transformer \emph{backbones} at similar width. We implement a lightweight diagonal SSM as an additional baseline; RiLM competes on \emph{decoding geometry} rather than on replacing $\varphi$ with a different recurrent operator.

\section{Method}
\label{sec:method}

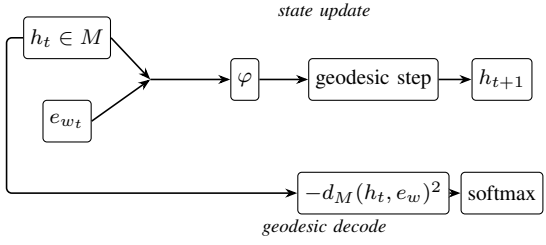
\begin{figure}[t]
\centering
\begin{tikzpicture}[
  font=\footnotesize,
  box/.style={draw, rounded corners=1.5pt, minimum height=5.5mm, inner sep=2.5pt, align=center},
  line/.style={-{Stealth[length=1.6mm]}, semithick},
  lbl/.style={font=\scriptsize\itshape}
]
\node[box] (ht) at (0, 0.55) {$h_t \in M$};
\node[box] (ew) at (0, -0.55) {$e_{w_t}$};
\coordinate (merge) at (1.1, 0);
\node[box] (phi) at (2.35, 0) {$\varphi$};
\node[box] (geo) at (4.05, 0) {geodesic step};
\node[box] (ht1) at (5.75, 0) {$h_{t+1}$};
\draw[line] (ht.east) -- (merge);
\draw[line] (ew.east) -- (merge);
\draw[line] (merge) -- (phi.west);
\draw[line] (phi) -- (geo);
\draw[line] (geo) -- (ht1);
\node[lbl] at (3.4, 0.9) {state update};

\node[box] (dist) at (4.05, -1.5) {$-d_M(h_t,e_w)^2$};
\node[box] (soft) at (5.75, -1.5) {softmax};
\draw[line, rounded corners=2pt]
  (ht.west) -- (-0.8, 0.55) -- (-0.8, -1.5) -- (dist.west);
\draw[line] (dist) -- (soft);
\node[lbl] at (3.4, -1.95) {geodesic decode};
\end{tikzpicture}
\caption{RiLM at timestep $t$: shared $\varphi$, geodesic state update, and distance-based decoding without $W_{\mathrm{out}}$.}
\label{fig:rilm}
\end{figure}

RiLM treats a sentence as motion on a manifold: each token nudges the state along a learned geodesic, and the next-word distribution is read off from geometry rather than a separate classifier. Figure~\ref{fig:rilm} summarizes one timestep. The design deliberately mirrors tied embeddings---input and output share parameters---but replaces the linear map $W_{\mathrm{out}} h$ with negative squared distance $-d_M(h,e_w)^2$, so the model must organize vocabulary points on $M$ in a way that supports both ingestion and prediction.

\subsection{Geometric Background}
A Riemannian manifold $(M,g)$ equips each tangent space $T_x M$ with an inner product $g_x$ varying smoothly in $x$. The exponential map $\mathrm{exp}_x \colon T_x M \to M$ sends tangent vector $v$ along the geodesic starting at $x$ in direction $v$; the logarithmic map $\log_x$ inverts it locally. Geodesic distance $d_M(x,y)$ is the length of the shortest path on $M$. On $\mathbb{R}^d$, $\mathrm{exp}_x(v)=x+v$ and $d_M(x,y)=\|x-y\|_2$.

The Poincar\'e ball $\mathbb{H}^d_c = \{x \in \mathbb{R}^d : \|x\| < 1/\sqrt{c}\}$ models hyperbolic space with curvature $c>0$. The conformal factor $\lambda_x^c = 2/(1-c\|x\|^2)$ diverges at the boundary, yielding exponential volume growth $\mathrm{Vol}(B_{\mathbb{H}}(r)) \approx e^{\sqrt{c}\,r}$. M\"obius addition~\cite{ganea2018},
\begin{equation}
x \oplus_c y = \frac{(1 + 2c\langle x, y\rangle + c\|y\|^2)x + (1 - c\|x\|^2)y}{1 + 2c\langle x, y\rangle + c^2\|x\|^2\|y\|^2},
\end{equation}
is closed on the open ball. Hyperbolic distance is $d_{\mathbb{H}}(x,y) = (2/\sqrt{c})\,\mathrm{arctanh}(\sqrt{c}\,\|(-x)\oplus_c y\|)$.

\subsection{Recurrence and Geodesic Decoding}
Given vocabulary $V$, each word $w$ has embedding $e_w \in M$ and the contextual state $h_t \in M$ summarizes prefix $w_{1:t}$. A shared composition map $\varphi \colon \mathbb{R}^d \times \mathbb{R}^d \to \mathbb{R}^d$ reads tangent coordinates at the origin and proposes an update direction:
\begin{equation}
\label{eq:recurrence}
h_{t+1} = \mathrm{exp}_{h_t}\!\bigl(\varphi(\log_0(h_t),\, \log_0(e_{w_t}))\bigr).
\end{equation}
Decoding with temperature $\tau > 0$ defines a softmax over \emph{negative squared distances},
\begin{equation}
\label{eq:decode}
p(w \mid h_t) = \frac{\exp(-d_M(h_t, e_w)^2 / \tau)}{\sum_{w'} \exp(-d_M(h_t, e_{w'})^2 / \tau)}.
\end{equation}
which is analogous to metric learning: likely words are those whose embeddings lie in a small geodesic ball around $h_t$. The softmax can be viewed as a kernel classifier with $k(x,y)=\exp(-d_M(x,y)^2/\tau)$; on $\mathbb{R}^d$ this resembles a Gaussian RBF over embedding space, while on $\mathbb{H}^d$ distances grow faster with separation, sharpening the effective decision boundary near the state trajectory. Unlike a free linear head $W_{\mathrm{out}} h$, the classifier weights are \emph{identified} with vocabulary geometry---the model cannot assign high probability to a word without placing $e_w$ near regions of $M$ that $h_t$ actually visits.

On $\mathbb{R}^d$, identifying tangent vectors with ambient coordinates ($\log_0(x){=}x$, $\mathrm{exp}_h(v){=}h{+}v$), Eq.~\eqref{eq:recurrence} reduces to additive composition $h_{t+1} = h_t + \varphi(h_t, e_{w_t})$---the Euclidean special case of our unpublished flat prototype (anonymous supplementary code). The sequence starts from $h_0 = \mathrm{exp}_0(\theta_{\mathrm{root}}) = \theta_{\mathrm{root}}$ with learned $\theta_{\mathrm{root}} \in \mathbb{R}^d$.

\subsection{Hyperbolic RiLM and M\"obius Recurrence}
Eq.~\eqref{eq:recurrence} is the natural hyperbolic analogue of additive recurrence, but implementing $\mathrm{exp}_{h_t}$ directly on $\mathbb{H}^d$ fails in practice. As $\|h_t\| \to 1/\sqrt{c}$, the conformal factor $\lambda_{h_t} = 2/(1-c\|h_t\|^2)$ diverges: small tangent steps at $h_t$ correspond to enormous boundary motion, states stick to the ball boundary, and $d_{\mathbb{H}}(h_t, e_w)$ becomes nearly constant for all $w$. We observed $\|h_t\| \approx 0.999$ within three steps and validation perplexity equal to $|V|$ (Section~\ref{sec:discussion}). 

Our fix computes increments where the metric is well-behaved---at the origin---and transports them with M\"obius addition:
\begin{equation}
\label{eq:mobius}
\delta_t = \mathrm{exp}_0\!\bigl(s \cdot \varphi(\log_0 h_t, \log_0 e_{w_t})\bigr), \quad
h_{t+1} = h_t \oplus_c \delta_t,
\end{equation}
with learnable step scale $s$ initialized near $1/\sqrt{d}$. After training on WikiText-2, $\|h_t\|$ on held-out prefixes typically lies in $[0.29, 0.71]$---well inside the ball (Figure~\ref{fig:geometry}). Embeddings and $\theta_{\mathrm{root}}$ are projected to $\|x\| \le 0.999/\sqrt{c}$; projecting $h_t$ after every step stalled optimization at ${\sim}130$ PPL, so we project only static parameters.

\subsection{Composition Function and Training}
We use a single MLP for $\varphi$ in all main experiments: $\varphi(v_h, v_e) = \tanh(W[v_h; v_e] + b)$ with $W \in \mathbb{R}^{d \times 2d}$ (${\sim}33$k parameters at $d{=}128$), shared across timesteps. Sharing $\varphi$ is central to the RiLM hypothesis: composition rules should not proliferate with sequence length the way layer-specific attention maps do. The model minimizes masked token negative log-likelihood,
\begin{equation}
\mathcal{L} = -\frac{1}{\sum_t m_t} \sum_{t} m_t \log p(w_{t+1} \mid h_t),
\end{equation}
with padding mask $m_t \in \{0,1\}$. Long compositions are trained with truncated BPTT ($k{=}8$): states detach every $k$ steps, limiting gradient depth through repeated $\varphi$ while matching the effective update frequency of windowed baseline training. At $k{=}8$ and context 64, each token still participates in eight compositional steps before gradients truncate---comparable to the receptive field of a shallow LSTM unrolled over the same window. Under this parameter budget, most predictive signal appears local: extending context to 256 with spline $\varphi$ regresses to $65.6$ PPL (Table~\ref{tab:abl}), so main experiments keep context 64 and MLP $\varphi$. Flat RiLM uses Adam at $\eta = 10^{-3}$; HypRiLM uses $\eta = 3 \times 10^{-3}$, curvature $c = 1.0$, embedding initialization $\mathcal{N}(0, (0.3/\sqrt{d})^2)$, and gradient clipping at norm $1.0$. All models train for 10 epochs with batch size 128 and context length 64 (Table~\ref{tab:hp}). Volume-preserving word flows $\Phi_w(h)=\mathrm{exp}_h(\varphi(\log_0 h,\log_0 e_w))$ are possible in principle~\cite{rezende2015}; our MLP $\varphi$ does not satisfy this, and spline variants (Table~\ref{tab:abl}) move perplexity only slightly---suggesting that gains come from decoding geometry and manifold choice rather than from entropy-preserving dynamics alone.

\subsection{Parameter Budget and Baseline Fairness}
At $d{=}128$ and $|V|{=}2000$, geodesic decoding eliminates the ${\sim}256$k parameters that LSTM and Transformer decoders allocate to $W_{\mathrm{out}}$ (Table~\ref{tab:params}). RiLM reinvests that capacity in full-width embeddings and $\varphi$. Fair comparison therefore requires two baseline regimes. \textbf{Tied} models at $d{=}128$ share input/output embeddings like RiLM but retain flat LSTM, attention, or SSM dynamics. \textbf{Matched} models equalize total parameters at ${\sim}290$k by shrinking hidden width to $d' \approx 82$--$97$, which weakens their recurrent core. Tied rows are the stronger control on decoding; matched rows isolate whether RiLM wins only because it keeps $d{=}128$.

\begin{table}[t]
\centering
\caption{Parameter allocation ($d{=}128$, $|V|{=}2000$).}
\label{tab:params}
\small
\begin{tabular}{@{}lrrr@{}}
\toprule
Component & RiLM & LSTM & Transformer \\
\midrule
Embeddings & 256k & 256k & 256k \\
Output head $W_{\mathrm{out}}$ & 0 & 256k & 256k \\
Composition core & 33k & 264k & 265k \\
\midrule
Total & 289k & 776k & 777k \\
\bottomrule
\end{tabular}
\end{table}

\begin{table}[t]
\centering
\caption{Shared hyperparameters.}
\label{tab:hp}
\small
\begin{tabular}{@{}ll@{}}
\toprule
Setting & Value \\
\midrule
Embedding $d$ / vocab.\ $|V|$ & 128 / 2000 (most frequent) \\
Context length / TBPTT $k$ & 64 / 8 \\
Batch size / epochs & 128 / 10 \\
Decoding temperature $\tau$ & 1.0 \\
Flat / LSTM / TX learning rate & $10^{-3}$ \\
HypRiLM learning rate & $3 \times 10^{-3}$ \\
SSM learning rate & $3 \times 10^{-4}$ \\
LSTM / Transformer & 2 layers; TX: 4 heads, FFN 256 \\
WT-2 seeds / PTB seeds & $\{42,\ldots,46\}$ / $\{42,43,44\}$ \\
\bottomrule
\end{tabular}
\end{table}

\section{Experiments}
\label{sec:experiments}

We designed experiments to answer three questions. \textbf{Q1:} Does geodesic decoding improve language modeling when decoder capacity is fairly counted against LSTM, Transformer, and SSM baselines? \textbf{Q2:} Holding $\varphi$ fixed, does hyperbolic curvature improve over flat space? \textbf{Q3:} Do relative conclusions survive a tenfold increase in $|V|$? We report validation perplexity $\exp(\mathcal{L})$ on WikiText-2 (WT-2; full train/validation splits, 2.1M training tokens) and the Zaremba Penn Treebank (PTB; 887k training tokens). Primary tables use the \textbf{2000 most frequent words} on both corpora. This protocol fixes $|V|$ so that $W_{\mathrm{out}}$ costs ${\sim}256$k parameters at $d{=}128$, making the trade-off in Table~\ref{tab:params} explicit. Absolute perplexities are \textbf{not} comparable to full-vocabulary leaderboard entries (Table~\ref{tab:lit}); our claims concern controlled comparisons at matched scale.

\textbf{Baselines and training protocol.} We compare against two-layer LSTM, two-layer causal Transformer (four heads, FFN width 256), and a selective diagonal SSM inspired by Mamba~\cite{gu2023mamba}---each in untied, tied~\cite{press2017}, and matched (${\sim}290$k) regimes. SSM uses Adam at $\eta = 3 \times 10^{-4}$ and reports the best validation checkpoint within 10 epochs (it overfits after ${\sim}4$ epochs on WT-2); Flat RiLM, LSTM, and Transformer use $\eta = 10^{-3}$; HypRiLM uses $3 \times 10^{-3}$. All non-SSM models use the final epoch checkpoint. WikiText-2 aggregates mean $\pm$ std over five seeds $\{42,\ldots,46\}$; PTB over three seeds $\{42,43,44\}$. Effect sizes on WT-2 are large: the HypRiLM--Flat gap (${\sim}33$ PPL) exceeds $100\times$ the HypRiLM standard deviation ($0.2$), and HypRiLM--SSM-tied gaps (${\sim}59$ PPL) exceed $15\times$ the SSM-tied std ($3.8$). Reported means $\pm$ std over seeds therefore suffice; formal significance tests add little beyond these intervals. All training uses one GPU; reproduction scripts and unit tests for hyperbolic primitives are in anonymous supplementary material.

\subsection{WikiText-2 at Controlled Vocabulary}
Table~\ref{tab:main} reports headline WT-2 results. HypRiLM reaches $54.2 \pm 0.2$ perplexity at ${\sim}289$k parameters---the best result by a wide margin, with standard deviation far smaller than the gaps to baselines. Flat RiLM attains $87.6 \pm 0.6$ with the \emph{same} $\varphi$ and parameter count, isolating curvature: hyperbolic space accounts for a ${\sim}38\%$ relative improvement on this corpus. Untied LSTM ($149.9 \pm 2.7$) and Transformer ($137.4 \pm 5.1$) use ${\sim}776$k parameters yet remain ${\sim}84$--$96$ PPL worse, indicating that adding $W_{\mathrm{out}}$ capacity does not close the gap when the compositional core stays flat and Euclidean.

\begin{table}[t]
\centering
\caption{WikiText-2 validation perplexity (five seeds, 2k vocab).}
\label{tab:main}
\small
\begin{tabular}{@{}lrr@{}}
\toprule
Model & PPL $\downarrow$ & Params \\
\midrule
HypRiLM (MLP $\varphi$) & \textbf{54.2 $\pm$ 0.2} & 289k \\
Flat RiLM (MLP $\varphi$) & 87.6 $\pm$ 0.6 & 289k \\
LSTM, untied & 149.9 $\pm$ 2.7 & 776k \\
Transformer, untied & 137.4 $\pm$ 5.1 & 777k \\
\bottomrule
\end{tabular}
\end{table}

\begin{table}[t]
\centering
\caption{Fair WikiText-2 comparison (five seeds $\{42,\ldots,46\}$; all models). \textbf{Primary controls:} tied and matched rows.}
\label{tab:fair}
\small
\begin{tabular}{@{}lrr@{}}
\toprule
Model & PPL $\downarrow$ & Params \\
\midrule
HypRiLM & \textbf{54.2 $\pm$ 0.2} & 289k \\
Flat RiLM & 87.6 $\pm$ 0.6 & 289k \\
LSTM-tied ($d{=}128$) & 117.9 $\pm$ 1.4 & 520k \\
LSTM-matched & 125.2 $\pm$ 6.8 & 273k \\
Transformer-tied ($d{=}128$) & 147.0 $\pm$ 1.0 & 521k \\
Transformer-matched & 142.8 $\pm$ 1.1 & 283k \\
SSM-tied ($d{=}128$) & 113.0 $\pm$ 3.8 & 374k \\
SSM-matched & 118.1 $\pm$ 0.7 & 271k \\
\bottomrule
\end{tabular}
\end{table}

Fair comparisons in Table~\ref{tab:fair} address whether RiLM wins only by tying or by parameter accounting. Against \textbf{tied} LSTM ($117.9 \pm 1.4$), Transformer-tied ($147.0 \pm 1.0$), and SSM-tied ($113.0 \pm 3.8$)---all at $d{=}128$ with shared input/output weights---HypRiLM still leads by 59--93 PPL while using fewer parameters than the LSTM and Transformer rows. The SSM-tied row is particularly informative: it is a modern recurrent decoder with selective gating, yet remains ${\sim}2\times$ worse than HypRiLM ($113.0$ vs.\ $54.2$). Against \textbf{matched} baselines at ${\sim}290$k total, hidden widths shrink to $d' \approx 82$--$97$; HypRiLM wins by 64--89 PPL while retaining $d{=}128$. No fair regime reverses the ranking.

On a small held-out subset (10k training / 500 validation tokens; seed 42), model ordering matches Table~\ref{tab:main} (HypRiLM 76.4, Flat RiLM 103.0, LSTM 189.0, Transformer 148.5 PPL). Ablating the M\"obius fix---using $\mathrm{exp}_{h_t}$ directly in Eq.~\eqref{eq:recurrence}---drives perplexity to the uniform baseline $|V|{=}2000$ (Table~\ref{tab:collapse}), confirming that trainable hyperbolic dynamics are prerequisite to the headline results.

\subsection{Vocabulary Scaling to $|V|{=}10{,}000$}
When $|V|$ grows tenfold, embedding tables dominate parameter count (${\sim}1.3$M for RiLM at $d{=}128$) and the original ${\sim}290$k matched baselines are no longer meaningful---shrinking $d$ to match budget would yield $d' < 40$. We therefore compare RiLM variants and tied baselines at full width with batch size 32 (Table~\ref{tab:vocab10k}). Absolute perplexity rises and is not comparable to Table~\ref{tab:main}; the question is whether \emph{ranking} persists.

Both RiLM variants remain far below SSM-tied ($708.3 \pm 9.0$): Flat RiLM at $341.8 \pm 0.5$, HypRiLM at $345.8 \pm 16.0$. Flat RiLM is marginally best on mean; HypRiLM is competitive but seed 44 is an outlier ($368.4$ PPL), inflating variance. We interpret this as evidence that (i)~geodesic decoding remains advantageous at larger $|V|$, but (ii)~hyperbolic curvature is not uniformly superior---on this scale, Euclidean geometry is slightly more stable. Tied LSTM and Transformer runs were unstable under the shared 10-epoch schedule (PPL $>1400$), highlighting optimization fragility of flat decoders at large $|V|$ under matched training budgets; SSM-tied is the reliable recurrent control here.

\subsection{Penn Treebank and Cross-Corpus Behavior}
PTB reverses the \emph{within-family} WT-2 ordering: Flat RiLM ($40.9 \pm 0.6$) beats HypRiLM ($69.8 \pm 0.5$) at identical ${\sim}289$k parameters (Table~\ref{tab:ptb}). Against fair baselines, both variants remain far below tied LSTM ($214.3 \pm 11.3$), Transformer-tied ($182.3 \pm 1.2$), and SSM-tied ($108.9 \pm 23.7$), as well as matched LSTM ($161.6 \pm 1.6$), Transformer ($171.4 \pm 4.1$), and SSM ($149.3 \pm 27.3$) controls. Geodesic decoding therefore transfers across corpora even when negative curvature is not the right inductive bias. PTB is smaller and stylistically homogeneous (newswire); WT-2 is noisier web text with broader topical structure. We report both corpora so readers can judge when to prefer Flat vs.\ HypRiLM rather than treating hyperbolicity as a universal upgrade.

\subsection{Ablations and Literature Context}
Table~\ref{tab:abl} disentangles architectural choices on WT-2 (seed 42). Replacing MLP $\varphi$ with monotone spline compose nodes (8 knots) improves Flat RiLM by ${\sim}4.4$ PPL but changes HypRiLM by only ${\sim}0.1$ PPL---smaller shifts than the ${\sim}34$ PPL gap between hyperbolic and flat MLP rows. Reducing curvature to $c{=}0.1$ degrades HypRiLM from $54.0$ to $71.1$ PPL, confirming that the ball's negative curvature is doing work, not merely adding parameters. Extending context to 256 with spline $\varphi$ yields $65.6$ PPL, so we retain context 64 and MLP $\varphi$ in all headline runs.

Table~\ref{tab:lit} places our seed-42 numbers beside published full-vocabulary models \emph{for context only}. Transformer-XL, GPT-2, and AWD-LSTM operate at 33M--257M parameters with extensive regularization; HypRiLM uses 2k words and ${\sim}289$k parameters. The numeric proximity of HypRiLM ($54.0$) to AWD-LSTM ($60.7$) is \textbf{not} a controlled comparison and should not be read as competitive SOTA; it illustrates how controlled-vocabulary perplexity can look deceptively strong relative to leaderboard entries.

\begin{table}[t]
\centering
\caption{WikiText-2 at 10k vocabulary ($d{=}128$, three seeds).}
\label{tab:vocab10k}
\small
\begin{tabular}{@{}lrr@{}}
\toprule
Model & PPL $\downarrow$ & Params \\
\midrule
HypRiLM & 345.8 $\pm$ 16.0 & ${\sim}1.31$M \\
Flat RiLM & \textbf{341.8 $\pm$ 0.5} & ${\sim}1.31$M \\
SSM-tied ($d{=}128$) & 708.3 $\pm$ 9.0 & ${\sim}1.38$M \\
LSTM-tied ($d{=}128$)$^\dagger$ & 2021 $\pm$ 135 & ${\sim}1.54$M \\
Transformer-tied ($d{=}128$)$^\dagger$ & 1513 $\pm$ 116 & ${\sim}1.52$M \\
\bottomrule
\multicolumn{3}{@{}l@{}}{\footnotesize $^\dagger$Unstable under identical training at $|V|{=}10$k.}
\end{tabular}
\end{table}

\begin{table}[t]
\centering
\caption{Penn Treebank validation perplexity (three seeds, 2k vocab; tied and matched baselines).}
\label{tab:ptb}
\small
\begin{tabular}{@{}lrr@{}}
\toprule
Model & PPL $\downarrow$ & Params \\
\midrule
Flat RiLM & \textbf{40.9 $\pm$ 0.6} & 289k \\
HypRiLM & 69.8 $\pm$ 0.5 & 289k \\
LSTM-tied ($d{=}128$) & 214.3 $\pm$ 11.3 & 520k \\
LSTM-matched & 161.6 $\pm$ 1.6 & 273k \\
Transformer-tied ($d{=}128$) & 182.3 $\pm$ 1.2 & 521k \\
Transformer-matched & 171.4 $\pm$ 4.1 & 283k \\
SSM-tied ($d{=}128$) & 108.9 $\pm$ 23.7 & 374k \\
SSM-matched & 149.3 $\pm$ 27.3 & 271k \\
\bottomrule
\end{tabular}
\end{table}

\begin{table}[t]
\centering
\caption{WikiText-2 ablations (seed 42).}
\label{tab:abl}
\small
\begin{tabular}{@{}llr@{}}
\toprule
Variant & Manifold & PPL $\downarrow$ \\
\midrule
MLP $\varphi$ (default) & $\mathbb{H}^d$, $c{=}1.0$ & 54.0 \\
MLP $\varphi$ (default) & $\mathbb{R}^d$ & 88.1 \\
Spline 1D (8 knots) & $\mathbb{H}^d$ & 53.9 \\
Spline 1D (8 knots) & $\mathbb{R}^d$ & 83.7 \\
MLP, $c{=}0.1$ & $\mathbb{H}^d$ & 71.1 \\
Spline $\varphi$, context 256 & $\mathbb{H}^d$ & 65.6 \\
\bottomrule
\end{tabular}
\end{table}

\begin{table}[t]
\centering
\caption{WikiText-2 perplexity in the literature (token-level; \textbf{not comparable} to our 2k-vocab setup).}
\label{tab:lit}
\small
\begin{tabular}{@{}lrr@{}}
\toprule
Model & PPL $\downarrow$ & Params \\
\midrule
Transformer-XL~\cite{dai2019} & 23.1 & 257M \\
GPT-2 small~\cite{radford2019} & 29.4 & 117M \\
AWD-LSTM~\cite{merity2018} & 60.7 & 33M \\
HypRiLM (ours, seed 42) & 54.0 & 289k \\
Flat RiLM (ours, seed 42) & 88.1 & 289k \\
\bottomrule
\end{tabular}
\end{table}

\begin{table}[t]
\centering
\caption{Boundary collapse vs.\ M\"obius fix (seed 42). Na\"ive $\mathrm{exp}_{h_t}$: 10k train / 500 val subset. M\"obius: $\|h_t\|$ after training; PPL on full WT-2 validation.}
\label{tab:collapse}
\small
\begin{tabular}{@{}lrr@{}}
\toprule
Recurrence & $\|h_t\|$ after 3 steps & PPL \\
\midrule
$\mathrm{exp}_{h_t}(\varphi)$ (na\"ive) & ${\approx}0.999$ & 2000 \\
M\"obius Eq.~\eqref{eq:mobius} & $0.29$--$0.71$ & ${\sim}54$ \\
\bottomrule
\end{tabular}
\end{table}

\subsection{Learned Geometry and Training Dynamics}
Quantitative perplexity alone does not show whether geodesic decoding organizes representation space. We therefore inspect trained HypRiLM and Flat RiLM models (WT-2, seed 42, 10 epochs) on a held-out validation prefix.

\begin{figure}[t]
\centering
\includegraphics[width=0.72\linewidth]{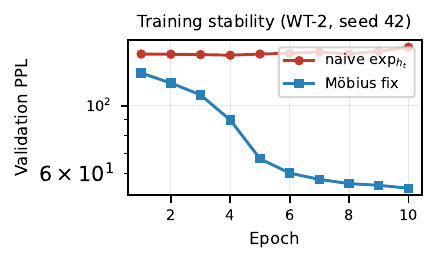}
\caption{Validation perplexity per epoch on WikiText-2 (seed 42): naive $\mathrm{exp}_{h_t}$ recurrence vs.\ M\"obius HypRiLM.}
\label{fig:train_stability}
\end{figure}

\textbf{Training stability.} Figure~\ref{fig:train_stability} tracks validation perplexity when the hyperbolic recurrence uses naive $\mathrm{exp}_{h_t}$ updates versus the M\"obius fix (Eq.~\eqref{eq:mobius}). Naive training remains near ${\sim}150$ PPL across 10 epochs; M\"obius HypRiLM improves monotonically to ${\sim}54$---the same regime as Table~\ref{tab:main}. The forward-pass collapse in Table~\ref{tab:collapse} ($\|h_t\|{\approx}0.999$, PPL $={}|V|$) explains \emph{why} naive training fails to descend; M\"obius transport is required for usable hyperbolic dynamics.

\begin{figure}[t]
\centering
\includegraphics[width=\linewidth]{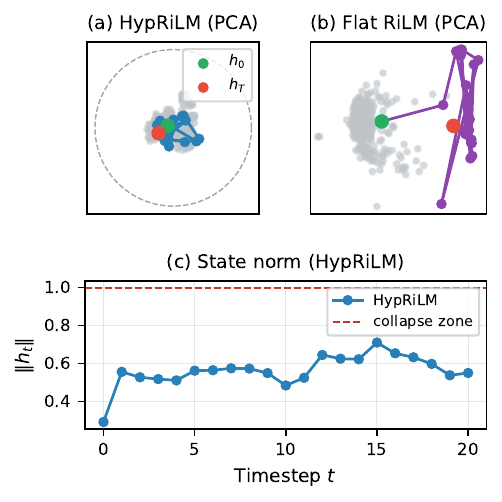}
\caption{Learned geometry on a WT-2 validation prefix (seed 42). (a)--(b)~PCA projections with Poincar\'e boundary (dashed) in (a); (c)~$\|h_t\|$ along the HypRiLM trajectory.}
\label{fig:geometry}
\end{figure}

\textbf{Trajectory and ball geometry.} Figure~\ref{fig:geometry} visualizes a 20-token prefix. Panels (a--b) project vocabulary embeddings (gray) and the state trajectory $h_0 \!\to\! \cdots \!\to\! h_T$ (colored path; green/red: start/end) via PCA; the dashed circle in (a) is the Poincar\'e boundary ($\|x\|{=}1$ at $c{=}1$). Along this prefix, HypRiLM satisfies $\|h_t\| \in [0.29, 0.71]$ (panel c)---inside the ball and consistent with the post-training range in Table~\ref{tab:collapse}. Flat RiLM follows a comparable low-dimensional path in $\mathbb{R}^d$.

\textbf{Distance-based decoding in practice.} After the prefix \emph{``along with $\cdots$ city of''}, both models rank \texttt{was} and related function words among the highest-probability continuations (Table~\ref{tab:top5}). Neither model was cherry-picked on this example; it is the first long validation prefix returned by our data loader.

\begin{table}[t]
\centering
\caption{Top-5 next-token candidates by geodesic distance after prefix \emph{``along with $\cdots$ city of''} (seed 42; logits are $-d_M^2/\tau$).}
\label{tab:top5}
\small
\begin{tabular}{@{}ll@{}}
\toprule
Rank & HypRiLM / Flat RiLM \\
\midrule
1 & \texttt{<unk>} / \texttt{<unk>} \\
2 & \texttt{was} / \texttt{was} \\
3 & \texttt{is} / \texttt{and} \\
4 & \texttt{had} / \texttt{the} \\
5 & \texttt{the} / \texttt{in} \\
\bottomrule
\end{tabular}
\end{table}

\section{Discussion}
\label{sec:discussion}

\subsection{Boundary Collapse as a Trainability Phenomenon}
Our first HypRiLM implementation applied $\mathrm{exp}_{h_t}(\varphi(\cdot))$ directly in Eq.~\eqref{eq:recurrence}. The hyperbolic exponential map can be written $\mathrm{exp}_{h_t}(v) = h_t \oplus_c \mathrm{exp}_0(\tanh(\sqrt{c}\,\lambda_{h_t}\,\|v\|/2)\,v/(\sqrt{c}\,\|v\|))$. As $\|h_t\| \to 1/\sqrt{c}$, the conformal factor $\lambda_{h_t} = 2/(1-c\|h_t\|^2)$ grows without bound, $\tanh$ saturates, and updated states remain at the boundary. Within three recurrence steps we observed $\|h_t\| \approx 0.999$; geodesic distances to all vocabulary embeddings became nearly equal; logits flattened; and perplexity reached $|V| = 2000$---exactly the uniform-guess baseline (Table~\ref{tab:collapse}). This \emph{boundary collapse} is not a small optimization issue: it renders naive hyperbolic recurrence untrainable for language modeling at modest depth.

Equation~\eqref{eq:mobius} sidesteps the pathology by computing tangent increments at the origin---where $\lambda_0$ is moderate---and translating via $\oplus_c$, which maps interior points to interior points. After training, $\|h_t\|$ stays in $[0.29, 0.71]$ on held-out prefixes and perplexity falls to ${\sim}54$. Hard projection of $h_t$ after every step was counterproductive (${\sim}130$ PPL), likely because it interrupts gradient flow along geodesics; we project only embeddings and $\theta_{\mathrm{root}}$. We regard stable M\"obius recurrence as a \emph{necessary} engineering contribution alongside geodesic decoding: without it, the hyperbolic instantiation of RiLM does not function.

\subsection{Why Geodesic Decoding Helps at Small Scale}
When $|V|$ and $d$ are moderate, $W_{\mathrm{out}}$ can consume roughly one third of a standard untied LM's parameters (Table~\ref{tab:params}). Matched baselines respond by shrinking hidden width, which weakens LSTM gates or attention heads. RiLM instead keeps $d{=}128$ and forces the vocabulary embedding table to serve double duty: tokens are inputs \emph{and} the targets of distance-based classification. The model cannot learn arbitrary linear functionals of $h_t$ independent of where words live on $M$; plausible continuations must lie near the contextual trajectory. This coupling may explain why tied LSTM/Transformer models---which remove $W_{\mathrm{out}}$ but decode linearly in $\mathbb{R}^d$---still trail by tens of perplexity points.

Swapping $\mathbb{R}^d$ for $\mathbb{H}^d$ at fixed $\varphi$ yields a large WT-2 gain but not a PTB gain. We do not claim a formal hierarchy theorem for web text; rather, WT-2's broader topical and lexical structure may benefit from the exponential separation hyperbolic space affords, whereas compact PTB sentences are adequately served by flat geometry. Table~\ref{tab:guide} summarizes when each instantiation is preferable in our experiments.

\begin{table}[t]
\centering
\caption{Practical guide: which RiLM variant to try first (under our ${\sim}290$k, 2k-vocab protocol).}
\label{tab:guide}
\small
\begin{tabular}{@{}p{0.42\linewidth}p{0.46\linewidth}@{}}
\toprule
Setting & Recommendation \\
\midrule
Noisy web text (WT-2) & HypRiLM (large margin over flat and baselines) \\
Compact newswire (PTB) & Flat RiLM (beats HypRiLM at same $\varphi$) \\
$|V|$ $\gtrsim$ 10k, fixed training budget & Flat RiLM (more stable; similar mean PPL) \\
Need fastest inference & Flat RiLM (${\sim}0.15$ vs.\ ${\sim}0.52$ ms/token) \\
\bottomrule
\end{tabular}
\end{table}

Practitioners should treat manifold choice as a validation-set decision, not a default.

\subsection{Relation to Tied Embeddings}
Tied LSTM/Transformer models set $W_{\mathrm{out}} = E^\top$ where $E \in \mathbb{R}^{|V| \times d}$ is the embedding table, so logits are $E h_t$---a bilinear form in $\mathbb{R}^d$. RiLM instead uses $-\|h_t - e_w\|^2$ on $\mathbb{R}^d$ (or $-d_{\mathbb{H}}(h_t,e_w)^2$ on $\mathbb{H}^d$). Expanding the Euclidean case,
$-\|h - e_w\|^2 = -\|h\|^2 + 2\langle h, e_w \rangle - \|e_w\|^2$.
The term $-\|h\|^2$ is shared across all words at a given timestep and cancels in the softmax; RiLM logits therefore differ from tied linear decoding by the word-dependent norms $\|e_w\|^2$ and by whether inner products or distances define similarity. Geodesic decoding is not equivalent to tying alone: it imposes a metric structure on how hidden states relate to \emph{all} vocabulary points simultaneously. Our tied baselines isolate this distinction---they remove $W_{\mathrm{out}}$ but keep linear scoring---and still trail HypRiLM by 59+ PPL on WT-2.

\subsection{Efficiency and Limitations}
Per generated token, RiLM requires $O(d)$ work for $\varphi$ and $O(|V| \cdot d)$ for distance evaluation---the same order as softmax over $|V|$. The benefit is \emph{parametric}: eliminating $W_{\mathrm{out}}$ at small $|V|$. Measured on WikiText-2 (seed 42, autoregressive ms/token), Flat RiLM averages 0.15\,ms/token, HypRiLM 0.52\,ms/token, LSTM 0.44\,ms/token, and Transformer 0.40\,ms/token. Hyperbolic operations add ${\sim}3.5\times$ overhead over Flat RiLM but remain in the same ballpark as the recurrent baselines; Flat RiLM is the fastest variant in this benchmark.

Our study is intentionally bounded. Results use ${<}2$M parameters in the largest 10k-vocab setting, context 64, and TBPTT $k{=}8$. The 2k vocabulary is a controlled instrument, not a claim about full WikiText-2 benchmarking. Our SSM baseline is a lightweight diagonal model, not a tuned Mamba implementation; SSM uses best-validation checkpoint selection while other models use the final epoch---a potential advantage for SSM that our headline gaps absorb. Matched baselines equalize parameter count but not representation width. Reported ms/token timings use autoregressive generation with model-specific prefix lengths (see supplementary material) and are indicative, not hardware-independent benchmarks. Product-manifold variants improved WT-2 by only ${\sim}3$ PPL at higher complexity in exploratory runs. All models are trained from scratch without distillation.

\textbf{Scaling outlook.} At full vocabulary ($|V|{\approx}33$k), $W_{\mathrm{out}}$ dominates even billion-parameter transformers---but RiLM's $O(|V| \cdot d)$ decoding cost remains, and geodesic distance over the full table may need approximation (hierarchical softmax, sampled negatives, or product-manifold factorization of $V$). Our 10k study is a first stress test: RiLM variants stay ${\sim}2\times$ better than SSM-tied but flat geometry becomes competitive with hyperbolic. We expect the \emph{decoding} idea to matter most where $|V|$ is moderate and parameters scarce; whether curvature helps at scale is an open empirical question we do not resolve here. Extending RiLM to full vocabulary and wider embeddings is the natural next step for community-scale evaluation.

\section{Conclusion}
\label{sec:conclusion}

RiLM reframes small language modeling as geometry: context is a path on a manifold, and decoding reads distances to vocabulary points rather than applying a separate output matrix. With a shared MLP $\varphi$ and ${\sim}290$k parameters, HypRiLM leads on WikiText-2 against fair LSTM, Transformer, and SSM baselines; Flat RiLM is preferable on Penn Treebank; both beat tied and matched controls on both corpora. Stable M\"obius recurrence proved essential---without it, hyperbolic RiLM does not train.

Within the controlled regime we study, geodesic decoding is a concrete way to couple representation and prediction when parameters are scarce. It is not a drop-in replacement for billion-parameter pretraining, but it offers a principled alternative to $W_{\mathrm{out}}$ whose benefits persist under tied and matched comparisons and partially under larger $|V|$. We hope the framework, the boundary-collapse analysis, and the multi-seed protocol provide a reproducible baseline for future work on geometric language models.

\bibliographystyle{IEEEtran}

\end{document}